\documentclass{article} 
\usepackage{iclr2027_conference,times}

\usepackage{amsmath,amsfonts,bm}

\def\eqref#1{equation~\ref{#1}}

\def\1{\bm{1}}

\DeclareMathAlphabet{\mathsfit}{\encodingdefault}{\sfdefault}{m}{sl}
\SetMathAlphabet{\mathsfit}{bold}{\encodingdefault}{\sfdefault}{bx}{n}

\usepackage{hyperref}
\usepackage{url}
\usepackage{graphicx}
\usepackage{booktabs}
\usepackage{multirow}
\usepackage{caption}
\usepackage[table]{xcolor}
\usepackage[T1]{fontenc}
\usepackage[most]{tcolorbox}

\newtcolorbox{promptbox}[1][]{
  enhanced,
  colback=black!2,
  colframe=black!65,
  boxrule=0pt,
  leftrule=2pt,
  arc=0pt,
  left=9pt, right=9pt, top=6pt, bottom=6pt,
  breakable,
  #1
}
\def\oursName{{GazeME}}

\title{Amplify What You Gaze At: Target Saliency Boosting in Text-to-Image Generation}

\author{Shengqi Dang$^{1,2}$, Zhengxi Yu$^{1,2}$, Feilin Han$^{1,2}$, Xingyu Lan$^{2,3}$, Nan Cao$^{1,2}$\thanks{ corresponding author} \\
$^1$Tongji University\\ 
$^2$Shanghai Innovation Institute\\
$^3$Fudan University \\
}

\iclrfinalcopy 
\begin{document}

\maketitle

\begin{abstract}
Text-to-image generation has advanced in controlling what, where, and how objects appear, yet how visual attention is distributed among objects remains largely unexplored. In this paper, we introduce \textit{Target Saliency Boosting}, a new task aimed at boosting the visual saliency of a specific object during text-to-image generation without requiring any visual priors. Our key insight is that visual saliency is inherently relative: boosting the saliency of a target object also depends on the global saliency distribution across all objects in the scene. Based on this insight, we propose~\oursName, a lightweight framework that uses saliency-marked prompts, inserting learnable marker tokens around object descriptions to indicate which objects to visually emphasize or suppress. To learn these markers, we construct a saliency-semantics dataset that associates objects in image--prompt pairs with object-level saliency scores, and propose Saliency Prior Marker Activation (SPMA), a saliency-aware stochastic marker activation strategy that exploits relative saliency relationships for robust training. During inference, \oursName~automatically inserts appropriate markers into the prompt, thereby directly enhancing the visual saliency of the target object. Extensive experiments demonstrate that \oursName~effectively boosts target saliency while preserving both semantic alignment and image quality.
\end{abstract}

\section{Introduction}

Text-to-image generation models have achieved remarkable success in synthesizing high-fidelity images from text prompts, empowering creative workflows across design, advertising, and the visual arts. Yet semantic fidelity marks only the beginning of effective visual communication. For designers, the decisive question is not simply what appears in the frame, but what seizes the viewer's attention first~\citep{Sutcliffe2008Getting}. An advertisement must make its product stand out, and a story illustration must direct focus to its protagonist. This shifts the challenge beyond semantics to the dynamics of visual attention: semantics dictates what to generate, whereas attention orchestrates where the eye is drawn, which is important for memory retention~\citep{santangelo2015forced}, cognitive comprehension~\citep{de2010attention}, and emotional resonance~\citep{subramanian2014emotion}.

Visual saliency provides a standard approach for modeling human visual attention, using a saliency map to quantify the spatial distribution of human attention over an image~\citep{kumain2025revisited}. While prior work has largely focused on predicting fixations from given images~\citep{zhao2026attend,kummerer2016deepgaze,kroner2020contextual}, the inverse problem, generating images in which specified targets become visually salient, remains largely unexplored. A recent method, GazeFusion~\citep{gazefusion}, conditions text-to-image diffusion models on explicit saliency map inputs, aiming to generate images whose saliency distribution is consistent with the provided maps. However, requiring users to manually specify meaningful saliency maps before image generation places a burden on them. This calls for more intuitive interaction paradigms that allow users to achieve saliency-driven generation with minimal explicit input.

To address this, we introduce a task named \textit{Target Saliency Boosting} in text-to-image generation, which offers an intuitive way to interact and control visual attention without explicit visual priors. Given a text prompt in which a target object is explicitly described, the objective is to generate an image where the target-object region exhibits high visual saliency. However, achieving this goal presents several challenges.
(1) Visual attention is complex, entangled across low-level contrast, global composition, and high-level semantics. Thus, traditional control methods are often biased, lacking precision for effective modulation.
(2) Saliency is relative and context-dependent. Boosting a target may suppress other elements and distort composition, breaking scene coherence.
(3) A generalizable dataset that provides text prompts and corresponding object-level saliency scores is currently unavailable, which precludes effective supervision for our task.

In this work, we propose \oursName, a lightweight framework for this task. Our key insight is that visual saliency is inherently relative: boosting the prominence of a target object depends on the attention distribution across non-target entities. Based on this insight, we introduce the concept of {saliency-marked prompts}, where learnable marker tokens are inserted around object descriptions to explicitly indicate which objects should be visually emphasized or suppressed. To learn these marker embeddings, we construct an automatic saliency-semantics dataset that maps individual objects within image--prompt pairs to their relative visual saliency scores. We further propose saliency prior marker activation (SPMA), a saliency-aware stochastic marker activation strategy that exploits the relative saliency relationships among objects to construct diverse training samples. Only the marker embeddings are optimized, and this approach is lightweight and preserves the generalization of the pretrained model. During inference, given a source prompt and a target object, \oursName{} automatically inserts the appropriate markers into the prompt, enabling direct manipulation of the relative visual prominence of objects without modifying their underlying semantic content.

Additionally, we conduct extensive experiments to evaluate the effectiveness of \oursName. Quantitative and qualitative comparisons, shortcut analyses, and ablation studies demonstrate that our method can effectively enhance target saliency while avoiding trivial solutions such as blurring the background. Subjective evaluations from user studies further validate the empirical advantages of GazeME, demonstrating alignment with human perceptual preferences.

In summary, our contributions are threefold:

\begin{itemize}
\item We introduce \textit{Target Saliency Boosting}, a novel task formulation that enables users to control visual saliency in text-to-image generation through natural language, without requiring any explicit saliency maps. This defines a new interaction paradigm where the model directly translates linguistic intent into visual saliency.

\item We propose \oursName, a lightweight framework that introduces learnable marker tokens to encode the intent of enhancing or suppressing visual saliency. The framework integrates saliency-aware data construction, embedding learning, and marker-guided generation, with saliency prior marker activation (SPMA) enabling robust training and the saliency-guided generation strategy reinforcing the desired visual saliency during inference.

\item We construct a language-grounded multi-object saliency dataset, where objects described in a prompt are associated with their visual saliency rankings and scores. We also develop an automatic annotation pipeline that derives object-level saliency supervision from image--prompt pairs, enabling scalable learning of saliency-aware marker embeddings.


\end{itemize}

\section{Related Work}
In this section, we review related works in visual saliency and controllable text-to-image generation. 

\subsection{Visual Saliency}
Visual saliency, originally conceptualized in cognitive psychology, refers to the visual property that makes certain regions stand out from their surroundings and naturally draw human attention~\citep{itti2007visual}. We review it from three aspects: prediction methods, datasets, and applications.

\textbf{Saliency Prediction.} Saliency prediction aims to model where humans tend to look in a scene. This target is usually represented as a saliency map (e.g., a heatmap where pixel values encode the likelihood of attracting fixations), with brighter regions indicating higher visual priority~\citep{koch1985shifts}. To this end, early computational models rely on hand-crafted features, such as multi-scale center-surround contrast~\citep{itti1998model} and spectral residuals~\citep{hou2007saliency}. Deep learning methods then adopt convolutional neural networks~\citep{kummerer2016deepgaze} and Transformers~\citep{lou2022transalnet} to learn hierarchical representations, significantly improving prediction accuracy. Recently, unified foundation models like Attend-to-Anything Model (AAM)~\citep{zhao2026attend} consolidate multiple saliency tasks into a single framework. These advances provide reliable tools for estimating visual attention and lay the foundation for downstream applications.

\textbf{Saliency Datasets.} Existing saliency datasets are typically collected by presenting visual stimuli to multiple observers under free-viewing conditions and recording their gaze trajectories with eye trackers. Representative datasets include MIT1003~\citep{judd2009learning}, U-EYE~\citep{jiang2023ueyes}, CAT2000~\citep{Borji2015CAT2000-CVPR}, SalECI~\citep{Jiang_2022_CVPR}, SALICON~\citep{Jiang_2015_CVPR}, and OSIE~\citep{Xu2014OSIE}, covering natural images, web pages, and e-commerce content. 
However, existing datasets lack prompt-level grounding and sufficient scene diversity; although some provide annotations for salient regions or the most salient objects, they still offer insufficient supervision for low-saliency objects as negative samples.

\textbf{Saliency-Guided Applications.} Saliency information has been widely exploited in downstream vision tasks and practical applications. In salient object detection and salient object segmentation, saliency cues help identify and localize visually prominent objects~\citep{Borji2019SalientObjectDetection}. Saliency has also been used for image cropping~\citep{xu2021saliency}, and image retrieval~\citep{wei2019saliency}, as well as region-of-interest (ROI) coding for perceptually efficient video compression~\citep{zhu2018spatiotemporal}. Recent methods such as GazeFusion further incorporate saliency information to guide image generation~\citep{gazefusion}. Despite these advances, creating images with user-specified saliency remains challenging, both in terms of interaction difficulty and the ability of generative models to understand and faithfully translate saliency intent.

\subsection{Controllable Text-to-Image Generation}
Controllable text-to-image generation extends standard text-to-image models with additional mechanisms that allow users to specify and refine desired properties of the generated images. We review it from two aspects: control modalities and control mechanisms.

\textbf{Control Modalities.} Controllable text-to-image generation extends natural language prompts with additional conditions to constrain the generated content. Common modalities include semantic segmentation masks, edge maps, depth maps, human poses, color palettes, and layout sketches, providing explicit spatial or structural guidance~\citep{mou2023t2i,Zhang_2023_ICCV,huang2023composer}. These conditions are typically encoded by dedicated modules and injected into pre-trained diffusion models. For example, AnyControl~\citep{sun2024anycontrol} supports multiple spatial conditions within a unified framework; GLIGEN~\citep{Li_2023_CVPR} grounds textual descriptions to specified spatial regions through bounding boxes; and SpaText~\citep{Avrahami_2023_CVPR} constructs spatially aligned semantic embeddings to enforce object-level layout constraints. Such modalities enable users to explicitly control properties, but using them generally requires semantic or spatial intent.


\textbf{Control Mechanisms.} Controllable generation can be achieved through different intervention mechanisms, including condition injection, embedding adaptation, attention manipulation, feature modulation, and denoising-process control. Condition-based methods such as ControlNet~\citep{Zhang_2023_ICCV} and T2I-Adapter~\citep{mou2023t2i} introduce auxiliary modules for external signals, while Textual Inversion~\citep{gal2022image} learns token embeddings to encode new concepts and LoRA~\citep{hu2021lora} enables parameter-efficient adaptation. Attention- and feature-based methods directly modify cross-attention maps or intermediate representations for semantic and appearance control~\citep{hertz2022prompt}. Alternatively, recent methods intervene in the denoising process itself, such as NoiseCtrl~\citep{dai2025noisectrl} and CogBlender~\citep{dang2026cogblender}, which manipulate sampling or velocity trajectories during generation. Despite these diverse mechanisms, existing methods primarily control \textit{what} appears, \textit{where} it appears, or \textit{how} it looks, while leaving \textit{how visual attention is distributed among objects} largely unexplored.

Overall, existing saliency prediction models and controllable text-to-image methods are insufficient to support our task. Saliency models mainly predict attention from existing images without semantic grounding, while controllable text-to-image methods focus on semantic, spatial, or appearance control, with limited support for saliency-aware generation. To bridge this gap, we introduce \oursName, enabling users to boost the visual saliency of a target object through natural language prompting.

\section{Methodology}
\begin{figure}
    \centering
    \includegraphics[width=1.0\linewidth]{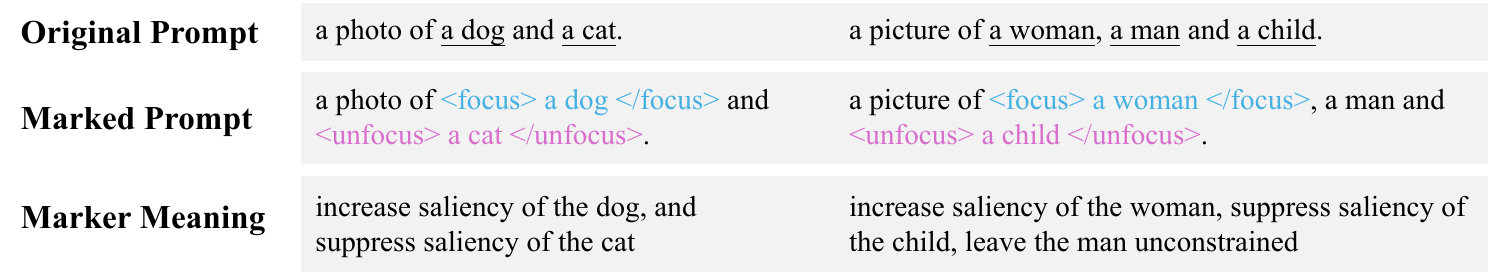}
    \caption{Examples of marked prompts, where \texttt{<focus></focus>} wraps the target object whose saliency should be boosted, while \texttt{<unfocus></unfocus>} wraps the object whose saliency should be suppressed.}
    \vspace{-0.2cm}
    \label{fig:marker}
\end{figure}

\begin{figure}
    \centering
    \includegraphics[width=\linewidth]{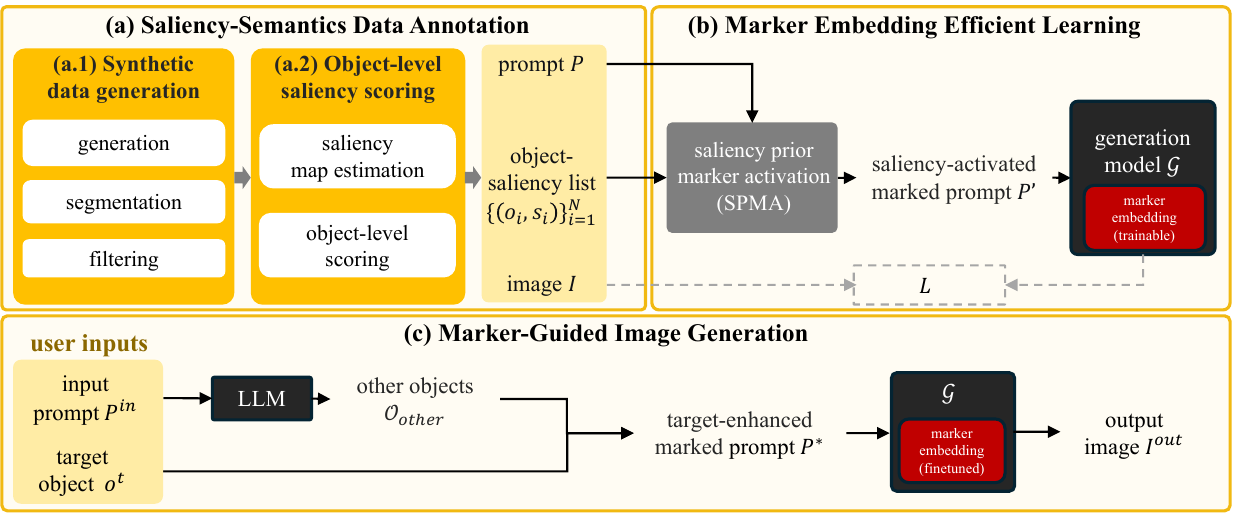}
    \caption{Overview of \oursName{}, which consists of three stages: (a) saliency-semantics data annotation, (b) marker embedding efficient learning, and (c) marker-guided image generation.}
    \vspace{-0.3cm}
    \label{fig:pipe}
\end{figure}

In this section, we present \oursName{} for \textit{Target Saliency Boosting}, centered on the concept of \textbf{saliency-marked prompts}. As shown in Figure~\ref{fig:marker}, we introduce two pairs of bidirectional marker tokens,  \texttt{<focus>}/\texttt{</focus>} and \texttt{<unfocus>}/\texttt{</unfocus>}, which are inserted around object descriptions to indicate whether an object should be visually emphasized or suppressed. \oursName{} learns saliency knowledge in these marker embeddings and uses them to guide image generation. As illustrated in Figure~\ref{fig:pipe}, our framework consists of three stages: (a)~\textbf{saliency-semantics data annotation}, which automatically infers object-level relative saliency from image--prompt pairs; (b)~\textbf{marker embedding efficient learning}, which optimizes only the marker embeddings to encode saliency knowledge; and (c)~\textbf{marker-guided image generation}, which inserts the learned markers into a user prompt \(P^{{in}}\) according to a target object \(o^t\) and generates the output image \(I^{{out}}\). We detail each stage below.

\subsection{Saliency-Semantics Data Annotation}
We construct a dataset of $1{,}637$ image--prompt pairs annotated with object-level saliency scores to train saliency-aware marker embeddings. Each sample is a tuple $\{I, P, \{(o_i, s_i)\}_{i=1}^{N}\}$, where $I$ is an image, $P$ is the corresponding prompt, and $s_i$ is the saliency score of the object $o_i$ mentioned in $P$, quantifying the average saliency within the object region relative to the global distribution. As shown in Figure~\ref{fig:pipe}, the data construction consists of two stages: (a.1) \textbf{synthetic data generation}, which produces diverse image--prompt pairs with object masks; and (a.2) \textbf{object-level saliency scoring}, which assigns a saliency score to each annotated object.

\textbf{Synthetic data generation.}
We obtain the data through three steps: data generation, segmentation, and filtering.
(1)~\textit{Generation.} We first prompt a Large Language Model (LLM, e.g., DeepSeek-V4-Flash in our implementation) to generate a diverse set of prompts, with each prompt \(P\) accompanied by a list of noun phrases \(\{o_1, o_2, \ldots, o_N\}\) denoting the objects in the scene. We then synthesize the corresponding image \(I\) using a text-to-image generation model (e.g., FLUX.1-dev in our implementation).
(2)~\textit{Segmentation.} Given \(I\), we obtain a binary object mask \(M_i\) for each object \(o_i\) using Segment Anything Model 3 (SAM3)~\citep{carion2025sam3segmentconcepts}, a state-of-the-art segmentation model capable of segmenting objects specified by natural-language concepts.
(3)~\textit{Filtering.} To ensure reliable object-level annotations, we remove samples with repeated object categories in the prompt, which may cause ambiguity in automatic object segmentation and matching. We also discard samples in which the queried object cannot be detected by SAM3.

\textbf{Object-level saliency scoring.}
We compute object-level saliency scores in two steps.
(1)~\textit{Saliency map estimation.} Given the synthesized image $I$, we compute a pixel-level saliency map $S$ using the Attend-to-Anything Model (AAM)~\citep{zhao2026attend}, a state-of-the-art foundation model for visual saliency prediction.
(2)~\textit{Object-level scoring.} Given the saliency map $S$ and the object mask $M_i$, we compute an object-level saliency score for each object by adapting the formulation of the normalized scanpath saliency (NSS)~\citep{bylinskii2018different}, defined as
\begin{equation}
s_i
=
\frac{1}{\sigma_{S}}
\left(
\frac{\sum_{p} S(p) M_i(p)}{|M_i|}
-
\mu_{S}
\right),
\label{eq:saliency-score}
\end{equation}
where $S(p)$ denotes the saliency value at pixel $p$, and $M_i(p) \in \{0,1\}$ indicates whether pixel $p$ belongs to the object mask $M_i$, $\mu_{S}$ and $\sigma_{S}$ denote the mean and the population standard deviation of $S$ over the whole image, respectively, and $|M_i|$ denotes the area of mask $M_i$ in pixels. The higher the value of $s_i$, the more salient the object region is; conversely, a lower value indicates lower saliency.

\subsection{Marker Embedding Efficient Learning}
\label{sec:marker-training}
With the constructed dataset, we train the saliency-aware marker embeddings to instill saliency awareness into the text-to-image model. To enhance robustness, generalization, and discriminative power along both positive and negative saliency directions, we propose \textbf{saliency prior marker activation} (SPMA), a lightweight, saliency score-conditioned mechanism that dynamically inserts markers into prompts during training. Moreover, instead of fine-tuning the full model, we freeze all backbone parameters and optimize only the newly introduced marker token embeddings. We next describe SPMA and the training procedure for learning the marker embeddings.

\textbf{Saliency prior marker activation (SPMA).}
SPMA takes a prompt \(P\) as input and dynamically outputs a saliency-activated marked prompt \(P'\) based on the saliency scores $\{s_i\}$ among objects $\{o_i\}$, forming diverse marker configurations that serve as an implicit ensemble over different saliency directions and object combinations. It encourages the marker embeddings to capture saliency relationships rather than memorize a fixed pattern. Specifically, SPMA consists of three steps: (1)~\textit{saliency-based grouping}, which divides objects into high- and low-saliency groups; (2)~\textit{stochastic marker activation}, which probabilistically selects objects to apply markers; and (3)~\textit{saliency-consistent truncation}, which applies a truncation operation to correct the activation values while preserving the relative saliency ordering among objects.

(1)~\textit{Saliency-based grouping.}
Given the saliency score $s_i$ of each object $o_i$, we first divide the objects into high- and low-saliency groups using a threshold $\gamma$:
\begin{equation}
y_i =
\begin{cases}
1, & s_i\geq\gamma,\\
0, & s_i<\gamma,
\end{cases}
\end{equation}
where \(y_i=1\) and \(y_i=0\) indicate high- and low-saliency objects, respectively. We set  \(\gamma\) as the per-sample mean saliency score to separate objects with positive and negative saliency score values.

(2)~\textit{Stochastic marker activation.}
We introduce a binary activation variable \(z_i \in \{0,1\}\) for each object \(o_i\) to indicate whether a marker is inserted around its description in the prompt, where \(z_i=1\) means that the object is selected and wrapped by its corresponding marker, and \(z_i=0\) means that no marker is inserted around it. The marker type is determined by the grouping label \(y_i\). Objects with \(y_i=1\) are wrapped by \texttt{<focus>}/\texttt{</focus>}, corresponding to high saliency, while objects with \(y_i=0\) are wrapped by \texttt{<unfocus>}/\texttt{</unfocus>}, corresponding to low saliency.

For each object, we sample $z_i$ from a Bernoulli distribution with saliency-dependent probability $p_i$. Specifically, we consider the following four cases.
For high-saliency objects ($y_i = 1$): the most salient object, i.e., the one with maximum $s_i$, is always activated with $p_i = 1$. For each remaining high-saliency object, its activation probability is defined by:
\begin{equation}
p_i = \frac{s_i - \gamma}{s_i^{\mathrm{up}} - \gamma},
\end{equation}
where $s_i^{\mathrm{up}} = \min_{s_j > s_i} s_j$ denotes the nearest higher saliency score.
For low-saliency objects ($y_i = 0$): the least salient object, i.e., the one with minimum $s_i$, is always activated with $p_i = 1$. For each remaining low-saliency object, its activation probability is defined by:
\begin{equation}
p_i = \frac{\gamma - s_i}{\gamma - s_i^{\mathrm{down}}},
\end{equation}
where $s_i^{\mathrm{down}} = \max_{s_j < s_i} s_j$ denotes the nearest lower saliency score.

(3)~\textit{Saliency-consistent truncation.}
Finally, we enforce saliency ordering consistency by truncating the sampled activation pattern $\{z_i\}$. Specifically, for the high-saliency group, an object can remain activated only if all objects with higher saliency scores are also activated; otherwise, it is deactivated. Conversely, for the low-saliency group, an object can remain activated only if all objects with lower saliency scores are also activated; otherwise, it is deactivated. 

This design has three key effects. Objects with a saliency score closer to \(\gamma\) are more likely to be deactivated, while objects farther from \(\gamma\) are more likely to be activated. Moreover, objects with similar saliency scores tend to be activated together. Additionally, the truncation rule preserves their relative saliency ordering, preventing contradictory activation patterns.

\textbf{Training procedure.}
We adopt a lightweight training strategy. 
With all other model parameters frozen, we optimize ${\theta}$ (the embeddings of four marker tokens: \texttt{<focus>}, \texttt{</focus>}, \texttt{<unfocus>}, and \texttt{</unfocus>}) using the flow matching objective:
\begin{equation}
    L
    =
    \mathbb{E}_{t,{\epsilon}}
    \left[
    \left\|
    v_{{\theta}}
    \left(
    {x}_t,t,P'
    \right)
    -
    {u}
    \right\|_2^2
    \right]
    \label{eq:loss}
\end{equation}
where ${x}_t$ denotes the latent representation of the original image $I$ after adding Gaussian noise (parameterized by $\epsilon$) at timestep $t$, ${u}$ denotes the target velocity, and $v_{{\theta}}({x}_t,t,{P'})$ denotes the velocity predicted by the generative model conditioned on the saliency-activated marked prompt $P'$.

\subsection{Marker-Guided Image Generation}

Based on the learned marker embeddings, we employ a saliency-enhanced generation strategy to explicitly enforce object-level saliency guidance during image generation. Given an input prompt \(P^{in}\) and a target object \(o^t\), we first employ an LLM to identify the other objects \(\mathcal{O}_{{other}}\) mentioned in \(P^{in}\). We then construct a saliency-enhanced marked prompt \(P^{*}\) by assigning the \texttt{<focus>} marker to the target object and the \texttt{<unfocus>} marker to the remaining objects. The resulting prompt explicitly represents \(o^t\) as the focus object and the objects in \(\mathcal{O}_{{other}}\) as unfocus objects. The constructed prompt \(P^{*}\) is then fed into the text-to-image generative model with the learned marker embeddings to synthesize the final image \(I^{{out}}\), thereby enhancing the visual saliency of the target object while suppressing the saliency of competing objects.

\textbf{Implementation details.}
We adopted FLUX.1-dev as the base text-to-image model. The model parameters were optimized using the AdamW optimizer with a learning rate of \(3{\times}10^{-4}\).Training was performed for 2,000 steps with a batch size of 2 on a single NVIDIA H200 GPU, taking approximately two hours. At inference, images were generated at \(512 \times 512\) resolution with 30 sampling steps. We set the random seed to 42 for all experiments.
\section{Experiments}
In this section, we first describe the experimental setup, then present the main comparison against baselines with shortcut analysis, followed by ablation studies. Additionally, we conduct a user study to validate the effectiveness of~\oursName{}.

\subsection{Experimental Setup}
In this section, we briefly introduce the experimental setup, including testset construction, baseline and metric selection. 

\textbf{Testset.}
We construct a target-saliency testset containing 432 challenging target saliency boosting tasks and each takes a prompt--target object pair as input. To build each pair, we generate diverse prompts, render images with the frozen base model, and select as the target an object that the untrained base model places in the bottom half of its own saliency score distribution (i.e., an object that is not salient by default). We will release the testset to ensure reproducibility.

\textbf{Baselines.}
We compare \oursName{} with five representative methods:
(1) \textit{Prompt Engineering}, which directly injects the target-saliency intent into the text-to-image prompt;
(2) \textit{Prompt Rewriting}, which queries an LLM to rewrite the input prompt into a more descriptive caption that emphasizes the target object;
(3) \textit{LoRA (full)}~\citep{hu2021lora}, which adopts low-rank adaptation to finetune the base model for target saliency enhancement, adapting all attention and feed-forward layers;
(4) \textit{LoRA (text)}, a capacity-localized LoRA variant that restricts adaptation to the text-side attention layers;
(5) \textit{FLUX-Editing}, which first generates an image and then edits it with an instruction to improve target saliency.

\textbf{Metrics.}
We evaluate all methods along three dimensions.
(1)~\textit{Saliency alignment.} We report Target-NSS, the saliency score of the target object, computed by Equation~\ref{eq:saliency-score}. Higher values indicate that the target object is more visually salient.
(2)~\textit{Semantic alignment.} We report CLIPScore~\citep{hessel2021clipscore}, the semantic similarity between the generated image and the input prompt. Higher values indicate better prompt fidelity.
(3)~\textit{Image quality.} We report CLIPIQA~\citep{wang2023exploring}, a no-reference perceptual quality metric. Higher values indicate better perceived image quality.

\begin{figure}
    \centering
    \includegraphics[width=1.0\linewidth]{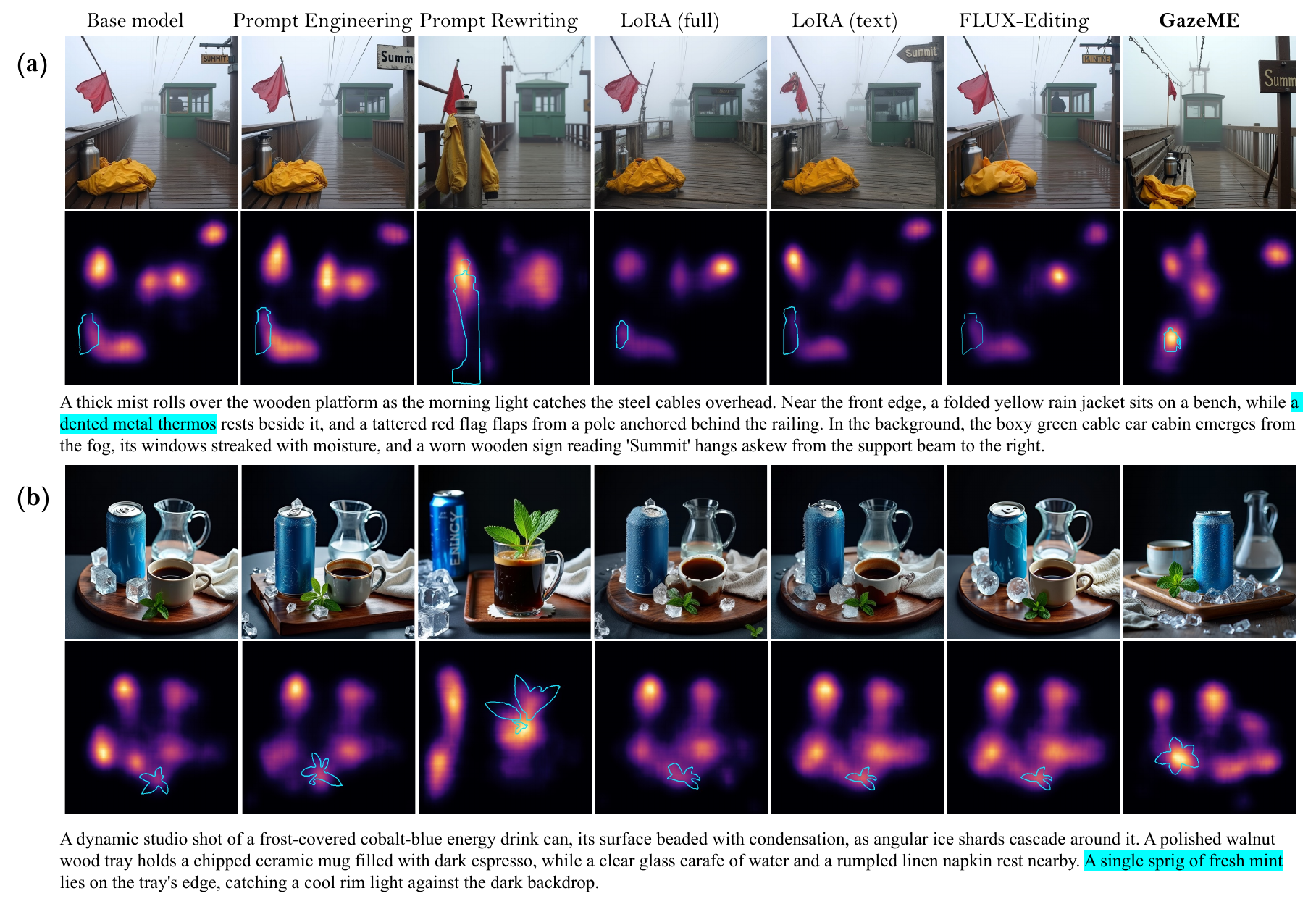}
    \caption{Qualitative results. We show the generated images and the corresponding saliency maps. The target object regions are outlined in cyan on the saliency maps. 
    }
    \label{fig:qualitative}
    \vspace{-0.3cm}
\end{figure}
\definecolor{bestblue}{RGB}{170,205,240}
\definecolor{secondblue}{RGB}{210,230,248}
\definecolor{thirdblue}{RGB}{235,245,252}

\begin{figure*}[t]
\centering
\footnotesize
\setlength{\tabcolsep}{3pt}

\begin{minipage}[c]{0.47\linewidth}
    \centering
    \captionof{table}{Quantitative comparison.}
    \label{tab:main-results}
    \resizebox{\linewidth}{!}{%
    \begin{tabular}{lccc}
    \toprule
    Method & Target-NSS$\uparrow$ & CLIPScore$\uparrow$ & CLIPIQA$\uparrow$ \\
    \midrule
    Base Model & 0.1791 & \cellcolor{thirdblue}0.2890 & 0.9463 \\
    \midrule
    Prompt Engineering & 0.2340 & 0.2846 & \cellcolor{bestblue}0.9523 \\
    Prompt Rewriting & 0.2378 & 0.2774 & \cellcolor{thirdblue}0.9485 \\
    \midrule
    LoRA (full) & \cellcolor{thirdblue}0.2521 & 0.2872 & 0.9429 \\
    LoRA (text) & 0.2313 & \cellcolor{secondblue}0.2896 & 0.9458 \\
    \midrule
    FLUX-Editing & \cellcolor{secondblue}0.2554 & \cellcolor{bestblue}0.2899 & 0.9426 \\
    \midrule
    \textbf{\oursName{}} & \cellcolor{bestblue}0.2761 & 0.2820 & \cellcolor{secondblue}0.9516 \\
    \bottomrule
    \end{tabular}%
    }
\end{minipage}
\hfill
\begin{minipage}[c]{0.51\linewidth}
    \centering
    \captionof{table}{Results of the ablation study.}
    \label{tab:ablations}
    \resizebox{\linewidth}{!}{%
    \begin{tabular}{lccc}
    \toprule
    Method & Target-NSS$\uparrow$ & CLIPScore$\uparrow$ & CLIPIQA$\uparrow$ \\
    \midrule
    \textbf{\oursName{}}($\gamma{=}\mu_{\text{sample}}$) & \cellcolor{bestblue}0.2761 & 0.2820 & \cellcolor{bestblue}0.9516 \\
    \midrule
    $\gamma{=}0$ & 0.2724 & 0.2819 & 0.9508 \\
    $\gamma{=}m_{\text{dataset}}$ & \cellcolor{thirdblue}0.2744 & \cellcolor{thirdblue}0.2821 & 0.9510 \\
    $\gamma{=}\mu_{\text{dataset}}$ & 0.2720 & 0.2820 & \cellcolor{secondblue}0.9515 \\
    $\gamma{=}m_{\text{sample}}$ & \cellcolor{secondblue}0.2757 & 0.2814 & \cellcolor{thirdblue}0.9512 \\
    \midrule
    w/o SPMA & 0.2276 & \cellcolor{secondblue}0.2864 & 0.9508 \\
    w/o training & 0.2192 & \cellcolor{bestblue}0.2870 & 0.9503  \\
    w/o \texttt{<unfocus>} in $P^*$ & 0.2521 & \cellcolor{secondblue}0.2864 & 0.9501 \\
    \bottomrule
    \end{tabular}%
    }
\end{minipage}

\end{figure*}

\subsection{Baseline Comparison}
We compare \oursName{} with the baselines from both quantitative and qualitative aspects. In addition, we perform a shortcut analysis with baselines on low-level factors that may act as saliency shortcuts.

\textbf{Quantitative Comparison.}
Table~\ref{tab:main-results} compares all methods on the testset. In terms of saliency alignment (Target-NSS), \oursName{} achieves the highest score and clearly outperforms prompt-level, training-based, and editing-based baselines. For semantic alignment (CLIPScore), \oursName{} remains competitive with the baselines and stays within a similar range, lying between Prompt Rewriting and Base Model, suggesting no substantial degradation in prompt fidelity. For visual quality (CLIPIQA), \oursName{} reaches the top level, matching the best baseline. Overall, these results demonstrate that \oursName{} is effective at enhancing target saliency while preserving semantic and perceptual quality, rather than merely optimizing a single metric.

\textbf{Qualitative Comparison.}
Figure~\ref{fig:qualitative} qualitatively compares \oursName{} with the baselines. \oursName{} consistently makes the target object the primary visual focus, whereas other methods, although capable of partially enhancing the target's saliency, often also amplify other objects or introduce unnatural artifacts, and thus fail to make the target stand out naturally. These examples are consistent with the quantitative results and further demonstrate that \oursName{} enhances target saliency in a natural and controllable manner.

\begin{figure*}[t]
\centering
\footnotesize
\setlength{\tabcolsep}{3pt}

\begin{minipage}[c]{0.48\linewidth}
    \centering
    \includegraphics[width=\linewidth]{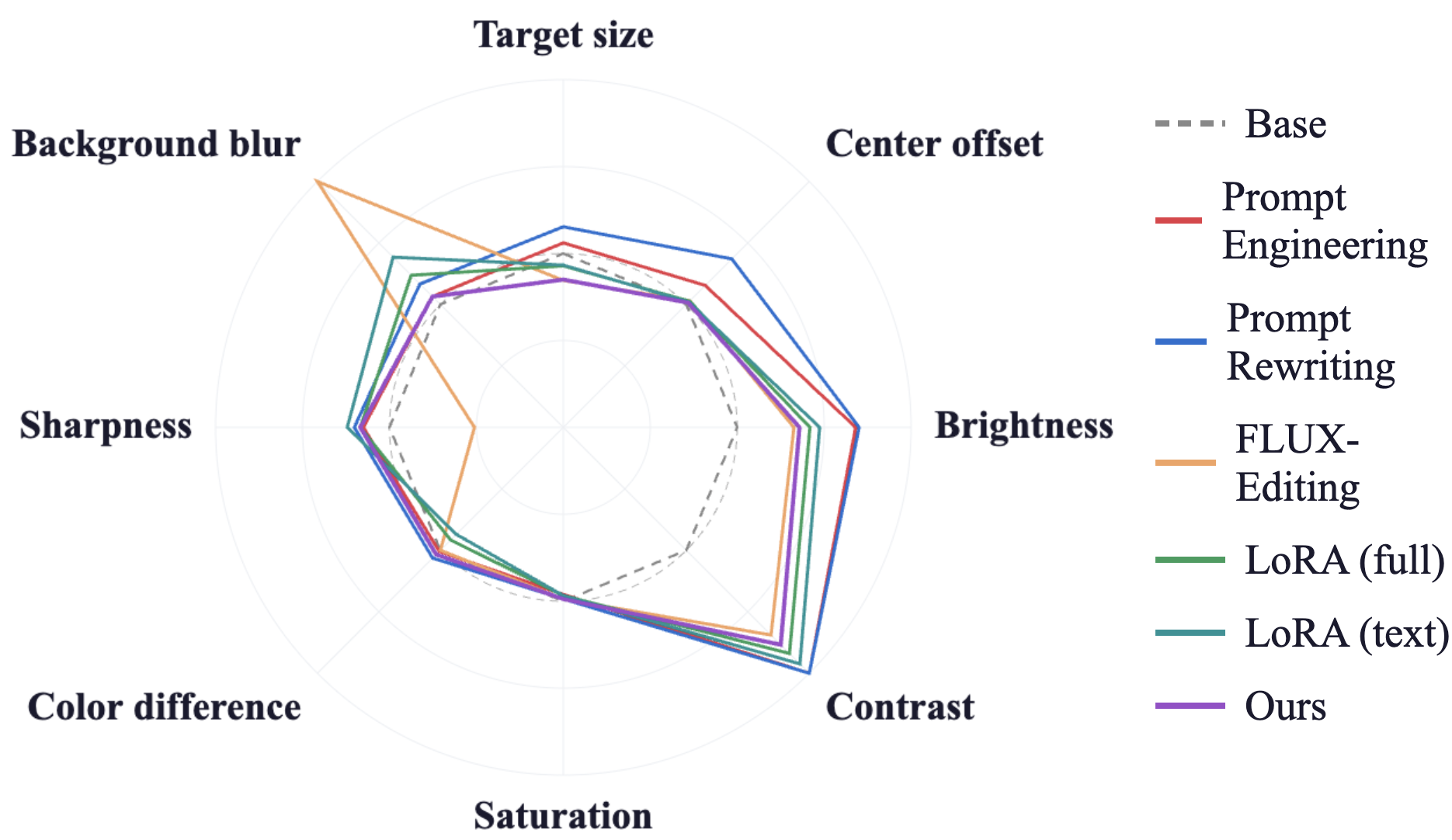}
    \captionsetup{type=figure, width=\linewidth, justification=centering, skip=2pt}
    \caption{Shortcut analysis.}
    \label{fig:shortcut-radar}
    
\end{minipage}
\hfill
\begin{minipage}[c]{0.47\linewidth}
    \centering
\captionsetup{type=figure, width=\linewidth, justification=centering, skip=2pt}
    \includegraphics[width=\linewidth]{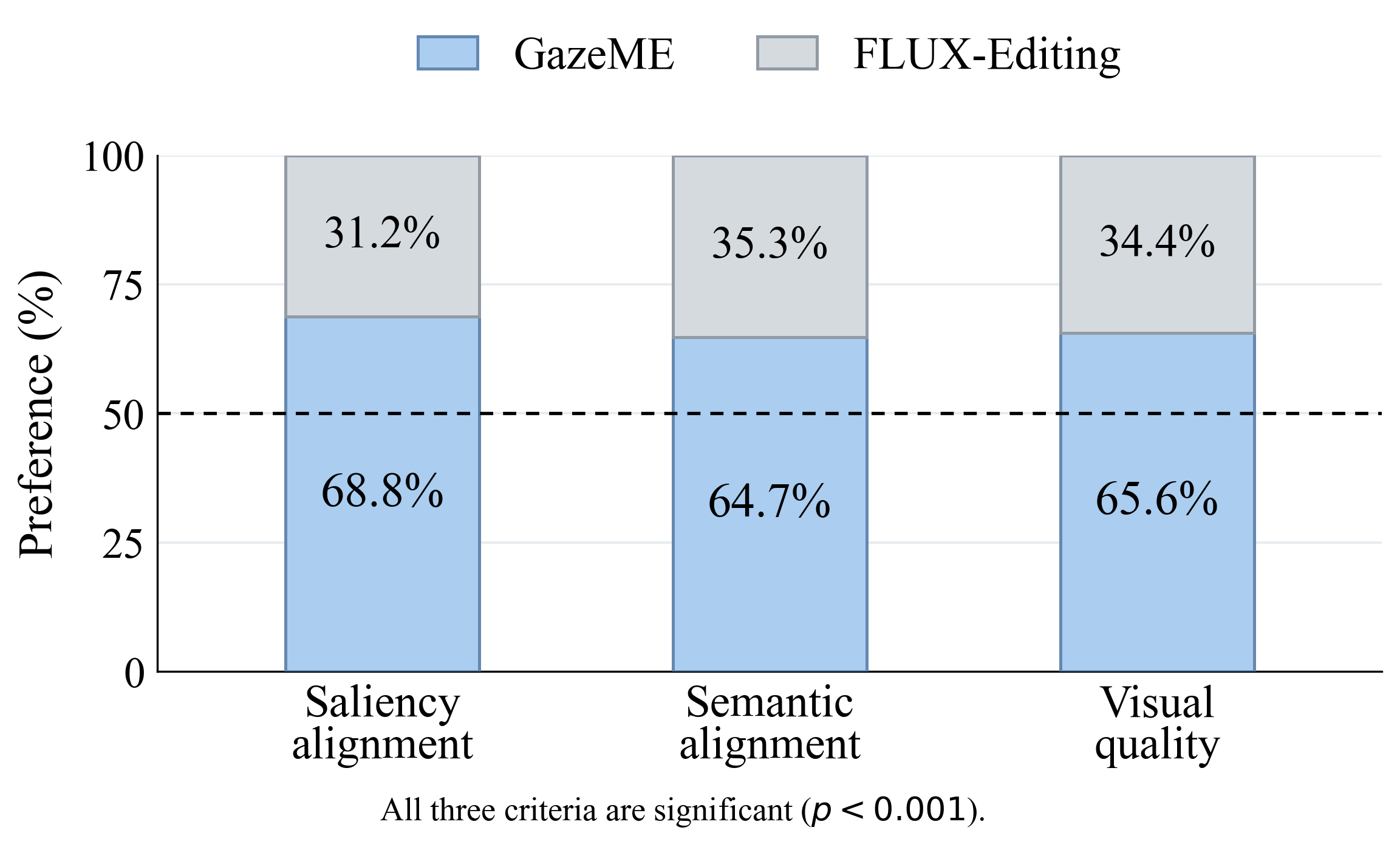}
    \captionsetup{type=figure, width=\linewidth, justification=centering, skip=2pt}
    \captionof{figure}{User study results.}
    \label{tab:userstudy}
\end{minipage}

\end{figure*}

\textbf{Shortcut Analysis.}
Saliency is not a direct function of low-level pixel statistics, yet low-level factors can mimic it, allowing methods to appear to improve saliency by biasing these factors rather than redistributing attention. To examine this, we analyze eight low-level visual factors, as shown in Figure~\ref{fig:shortcut-radar}. The radar chart (log scale) shows the ratio of each method's factor value to the base model's value (\(1.0\)) on each axis, except for center offset, where we plot the inverse ratio. The axes cover target size, center offset, brightness, contrast, saturation, color difference, sharpness, and background blur. FLUX-Editing mainly blurs the background, while prompt-level methods tend to enlarge the target, shift it toward the center, and increase brightness and contrast. \oursName{} shows deviations in brightness, and contrast, far smaller than other baselines and are not concentrated on a single axis. Overall, \oursName{} keeps the factor distribution closest to the base model, confirming that it improves saliency mainly through saliency redistribution rather than low-level factor shortcuts. 

\definecolor{bestblue}{RGB}{170,205,240}
\definecolor{secondblue}{RGB}{210,230,248}
\definecolor{thirdblue}{RGB}{235,245,252}

\subsection{Ablation Study}
We conduct ablation studies from two aspects of \oursName{}: the marker mechanism and the parameter selection. 

\textbf{Marker mechanism.}
We ablate three design choices of our marker mechanism: SPMA supervision, marker training, and the inclusion of the \texttt{<unfocus>} marker in $P^*$ (Table~\ref{tab:ablations}). Removing SPMA (\textit{w/o SPMA}, i.e., only the marker of the highest-saliency object is activated during training) drops Target-NSS from $0.2761$ to $0.2276$, and \textit{w/o training}, which keeps the marker embeddings fixed without optimization, drops it to $0.2192$, showing that both are essential for steering attention toward the target. Removing \texttt{<unfocus>} in $P^*$ reduces Target-NSS to $0.2521$, confirming that explicitly suppressing non-target objects further improves saliency control. 

\textbf{Parameter selection.}
We ablate the threshold $\gamma$ that splits objects into high- and low-saliency groups for SPMA. We compare a fixed threshold ($0$), dataset-level statistics ($m_{\text{dataset}}$ for the saliency score median and $\mu_{\text{dataset}}$ for the saliency score mean), and per-sample statistics ($m_{\text{sample}}$ and $\mu_{\text{sample}}$). As shown in Table~\ref{tab:ablations}, Target-NSS varies only slightly across all variants, while CLIPScore and CLIPIQA remain nearly constant. This indicates that \oursName{} is robust to the exact choice of $\gamma$. We adopt the per-sample mean setting ($\mu_{\text{sample}}$), which gives the best Target-NSS and guarantees that every sample contains both high- and low-saliency groups for bidirectional supervision.

\subsection{User Study}
\label{sec:userstudy}

We conducted a user study with {26 participants} to compare \oursName{} with the strongest baseline, \textit{FLUX-Editing}. We randomly selected 17 prompt--target object pairs as evaluation tasks, and for each task we generated an image pair consisting of one image from \oursName{} and one from the baseline. Each participant evaluated these 17 image pairs. The order of the two images within each pair was randomized. For each pair, participants answered three two-alternative questions to evaluate (1)~\textit{saliency alignment}, (2)~\textit{semantic alignment}, and (3)~\textit{visual quality}.

As shown in Figure~\ref{tab:userstudy}, \oursName{} received 68.8\% of the votes for saliency alignment, 64.7\% for semantic alignment, and 65.6\% for visual quality. We tested these participant-level majorities using two-sided exact binomial tests against a 50\% chance level. All three results remain significant ($p < 10^{-3}$), indicating a consistent preference for \oursName{} across the three criteria. The user study confirms the perceptual advantage of \oursName{} against the strongest baseline.

\section{Discussion}
Our results suggest that target saliency is best treated as a relative property. The bidirectional markers provide a lightweight interface for controlling this relativity. SPMA encourages the marker embeddings to learn robust saliency knowledge. 
Our work complements existing control over \textit{what} objects appear or \textit{where} they appear by targeting \textit{how attention is distributed} among them.

\textbf{Limitations and future work.}
Our approach has several limitations. First, we rely on automatic predictors for saliency estimation and object segmentation, which may introduce noise and bias into training. A promising solution is iterative refinement with human feedback, though this requires a careful annotation pipeline and a robust learning scheme for noisy or conflicting supervision. Second, while our markers encode relative saliency, we do not explicitly model spatial interactions or occlusion, which may limit performance in cluttered scenes. Extending the marker design to structured or pairwise relations is non-trivial due to their combinatorial nature. Finally, our framework targets static images; extending it to video would require temporally coherent markers to maintain saliency intent across frames, which is beyond our current scope.
\section{Conclusion}

In this paper, we introduce \textit{Target Saliency Boosting}, a new task for controlling visual attention in text-to-image generation without explicit visual priors. We propose \oursName{}, which encodes saliency knowledge into bidirectional marker tokens and applies them to guide image generation. Extensive experiments demonstrate that \oursName{} effectively boosts target saliency while preserving semantic alignment and image quality. We will release the constructed dataset and the source code to facilitate future research. We hope our work can inspire further exploration of visual attention-aware controllable generation and broader applications in areas such as computational advertising, visual design, and educational content production.

\bibliography{iclr2027_conference}

@article{santangelo2015forced,
title = {Forced to remember: When memory is biased by salient information},
journal = {Behavioural Brain Research},
volume = {283},
pages = {1-10},
year = {2015},

author = {Valerio Santangelo}
}

@article{de2010attention,
title = {Attention guiding in multimedia learning},
journal = {Learning and Instruction},
volume = {18},
number = {2},
pages = {135-145},
year = {2008},
author = {Eric Jamet and Monica Gavota and Christophe Quaireau}
}

@article{subramanian2014emotion,
    author = {Subramanian, Ramanathan and Shankar, Divya and Sebe, Nicu and Melcher, David},
    title = {Emotion modulates eye movement patterns and subsequent memory for the gist and details of movie scenes},
    journal = {Journal of Vision},
    volume = {14},
    number = {3},
    pages = {31-31},
    year = {2014},
    month = {03},
    
    eprint = {https://arvojournals.org/arvo/content_public/journal/jov/932817/i1534-7362-14-3-31.pdf},
}

@article{gazefusion,
author = {Zhang, Yunxiang and Wu, Nan and Lin, Connor Z. and Wetzstein, Gordon and Sun, Qi},
title = {GazeFusion: Saliency-Guided Image Generation},
year = {2024},
issue_date = {October 2024},
publisher = {Association for Computing Machinery},
address = {New York, NY, USA},
volume = {21},
number = {4},
journal = {ACM Transactions on Applied Perception},
month = nov,
articleno = {14},
numpages = {19}
}

@article{zhao2026attend,
  author       = {Wenzhuo Zhao and
                  Ronghao Xian and
                  Keren Fu and
                  Qijun Zhao},
  title        = {Attend to Anything: Foundation Model for Unified Human Attention Modeling},
  journal      = {CoRR},
  volume       = {abs/2606.03540},
  year         = {2026},
  eprinttype   = {arXiv},
  eprint       = {2606.03540},
  bibsource    = {dblp computer science bibliography, https://dblp.org}
}

@article{carion2025sam3segmentconcepts,
  author       = {Nicolas Carion and
                  Laura Gustafson and
                  Yuan{-}Ting Hu and
                  Shoubhik Debnath and
                  Ronghang Hu and
                  Didac Suris and
                  Chaitanya Ryali and
                  Kalyan Vasudev Alwala and
                  Haitham Khedr and
                  Andrew Huang and
                  Jie Lei and
                  Tengyu Ma and
                  Baishan Guo and
                  Arpit Kalla and
                  Markus Marks and
                  Joseph Greer and
                  Meng Wang and
                  Peize Sun and
                  Roman R{\"{a}}dle and
                  Triantafyllos Afouras and
                  Effrosyni Mavroudi and
                  Katherine Xu and
                  Tsung{-}Han Wu and
                  Yu Zhou and
                  Liliane Momeni and
                  Rishi Hazra and
                  Shuangrui Ding and
                  Sagar Vaze and
                  Francois Porcher and
                  Feng Li and
                  Siyuan Li and
                  Aishwarya Kamath and
                  Ho Kei Cheng and
                  Piotr Doll{\'{a}}r and
                  Nikhila Ravi and
                  Kate Saenko and
                  Pengchuan Zhang and
                  Christoph Feichtenhofer},
  title        = {{SAM} 3: Segment Anything with Concepts},
  journal      = {CoRR},
  volume       = {abs/2511.16719},
  year         = {2025},
  eprinttype   = {arXiv},
  eprint       = {2511.16719},
  bibsource    = {dblp computer science bibliography, https://dblp.org}
}

@ARTICLE{bylinskii2018different,
  author={Bylinskii, Zoya and Judd, Tilke and Oliva, Aude and Torralba, Antonio and Durand, Frédo},
  journal={IEEE Transactions on Pattern Analysis and Machine Intelligence}, 
  title={What Do Different Evaluation Metrics Tell Us About Saliency Models?}, 
  year={2019},
  volume={41},
  number={3},
  pages={740-757},
}

@inproceedings{Sutcliffe2008Getting,
author = {Sutcliffe, Alistair and Namoune, Abdallah},
title = {Getting the message across: visual attention, aesthetic design and what users remember},
year = {2008},
booktitle = {Proceedings of ACM Conference on Designing Interactive Systems},
pages = {11–20},
numpages = {10},
series = {DIS '08}
}

@article{kumain2025revisited,
author = {Kumain, Sandeep Chand and Singh, Maheep and Awasthi, Lalit Kumar},
title = {Revisited Visual Saliency Detection with Deep Learning: A Review of Recent Advancements},
year = {2025},
issue_date = {April 2026},
publisher = {Association for Computing Machinery},
address = {New York, NY, USA},
volume = {58},
number = {6},


journal = {ACM Computing Surveys},
month = dec,
articleno = {157},
numpages = {33}
}

@article{kummerer2016deepgaze,
  author       = {Matthias K{\"{u}}mmerer and
                  Thomas S. A. Wallis and
                  Matthias Bethge},
  title        = {DeepGaze {II:} Reading fixations from deep features trained on object
                  recognition},
  journal      = {CoRR},
  volume       = {abs/1610.01563},
  year         = {2016},
  eprinttype   = {arXiv},
  eprint       = {1610.01563},
  bibsource    = {dblp computer science bibliography, https://dblp.org}
}

@article{kroner2020contextual,
title = {Contextual encoder–decoder network for visual saliency prediction},
journal = {Neural Networks},
volume = {129},
pages = {261-270},
year = {2020},

author = {Alexander Kroner and Mario Senden and Kurt Driessens and Rainer Goebel},

}

@ARTICLE{itti1998model,
  author={Itti, L. and Koch, C. and Niebur, E.},
  journal={IEEE Transactions on Pattern Analysis and Machine Intelligence}, 
  title={A model of saliency-based visual attention for rapid scene analysis}, 
  year={1998},
  volume={20},
  number={11},
  pages={1254-1259},
  doi={10.1109/34.730558}}

@INPROCEEDINGS{hou2007saliency,
  author={Hou, Xiaodi and Zhang, Liqing},
  booktitle={Proceedings of IEEE Conference on Computer Vision and Pattern Recognition}, 
  title={Saliency Detection: A Spectral Residual Approach}, 
  year={2007},
  volume={},
  number={},
  pages={1-8},
  }

@article{itti2007visual,
  title={Visual salience},
  author={Itti, Laurent},
  journal={Scholarpedia},
  volume={2},
  number={9},
  pages={3327},
  year={2007}
}

@article{koch1985shifts,
  title={Shifts in selective visual attention: towards the underlying neural circuitry.},
  author={Koch, Christof and Ullman, Shimon},
  journal={Human neurobiology},
  volume={4},
  number={4},
  pages={219--227},
  year={1985}
}

@article{lou2022transalnet,
title = {TranSalNet: Towards perceptually relevant visual saliency prediction},
journal = {Neurocomputing},
volume = {494},
pages = {455-467},
year = {2022},

author = {Jianxun Lou and Hanhe Lin and David Marshall and Dietmar Saupe and Hantao Liu},

}

@INPROCEEDINGS{judd2009learning,
  author={Judd, Tilke and Ehinger, Krista and Durand, Frédo and Torralba, Antonio},
  booktitle={Proceedings of International Conference on Computer Vision}, 
  title={Learning to predict where humans look}, 
  year={2009},
  volume={},
  number={},
  pages={2106-2113},
  }

@article{Borji2015CAT2000-CVPR,
  author       = {Ali Borji and
                  Laurent Itti},
  title        = {{CAT2000:} {A} Large Scale Fixation Dataset for Boosting Saliency
                  Research},
  journal      = {CoRR},
  volume       = {abs/1505.03581},
  year         = {2015},
  eprinttype   = {arXiv},
  eprint       = {1505.03581},
  bibsource    = {dblp computer science bibliography, https://dblp.org}
}

@INPROCEEDINGS{Jiang_2022_CVPR,
  author={Jiang, Lai and Li, Yifei and Li, Shengxi and Xu, Mai and Lei, Se and Guo, Yichen and Huang, Bo},
  booktitle={Proceedings of IEEE/CVF Conference on Computer Vision and Pattern Recognition}, 
  title={Does text attract attention on e-commerce images: A novel saliency prediction dataset and method}, 
  year={2022},
  volume={},
  number={},
  pages={2078-2087},
}

@article{Xu2014OSIE,
    author = {Xu, Juan and Jiang, Ming and Wang, Shuo and Kankanhalli, Mohan S. and Zhao, Qi},
    title = {Predicting human gaze beyond pixels},
    journal = {Journal of Vision},
    volume = {14},
    number = {1},
    pages = {28-28},
    year = {2014},
    month = {01},
    
    
    eprint = {https://arvojournals.org/arvo/content_public/journal/jov/933546/i1534-7362-14-1-28.pdf},
}

@inproceedings{jiang2023ueyes,
author = {Jiang, Yue and Leiva, Luis A. and Rezazadegan Tavakoli, Hamed and R. B. Houssel, Paul and Kylm{\"a}l{\"a}, Julia and Oulasvirta, Antti},
title = {UEyes: Understanding Visual Saliency across User Interface Types},
year = {2023},

booktitle = {Proceedings of the SIGCHI Conference on Human Factors in Computing Systems},
articleno = {285},
numpages = {21},
}

@INPROCEEDINGS{Jiang_2015_CVPR,
  author={Jiang, Ming and Huang, Shengsheng and Duan, Juanyong and Zhao, Qi},
  booktitle={Proceedings of IEEE Conference on Computer Vision and Pattern Recognition}, 
  title={SALICON: Saliency in Context}, 
  year={2015},
  volume={},
  number={},
  pages={1072-1080}}

@article{Borji2019SalientObjectDetection,
  title={Salient object detection: A survey},
  author={Borji, Ali and Cheng, Ming-Ming and Hou, Qibin and Jiang, Huaizu and Li, Jia},
  journal={Computational Visual Media},
  volume={5},
  number={2},
  pages={117--150},
  year={2019}
}

@article{xu2021saliency,
title = {Saliency aware image cropping with latent region pair},
journal = {Expert Systems with Applications},
volume = {171},
pages = {114596},
year = {2021},

author = {Yifei Xu and Wujiang Xu and Mian Wang and Li Li and Genan Sang and Pingping Wei and Li Zhu},
 
}

@ARTICLE{wei2019saliency,
  author={Wei, Shikui and Liao, Lixin and Li, Jia and Zheng, Qinjie and Yang, Fei and Zhao, Yao},
  journal={IEEE Transactions on Image Processing}, 
  title={Saliency Inside: Learning Attentive CNNs for Content-Based Image Retrieval}, 
  year={2019},
  volume={28},
  number={9},
  pages={4580-4593},
}

@article{zhu2018spatiotemporal,
title = {Spatiotemporal visual saliency guided perceptual high efficiency video coding with neural network},
journal = {Neurocomputing},
volume = {275},
pages = {511-522},
year = {2018},

author = {Shiping Zhu and Ziyao Xu},

}

@inproceedings{sun2024anycontrol,
  author       = {Yanan Sun and
                  Yanchen Liu and
                  Yinhao Tang and
                  Wenjie Pei and
                  Kai Chen},
  title        = {AnyControl: Create Your Artwork with Versatile Control on Text-to-Image
                  Generation},
  booktitle    = {Proceedings of European Conference on Computer Vision},
  volume       = {15069},
  pages        = {92--109},
  year         = {2024},
  
  bibsource    = {dblp computer science bibliography, https://dblp.org}
}

@inproceedings{Li_2023_CVPR,
  author       = {Yuheng Li and
                  Haotian Liu and
                  Qingyang Wu and
                  Fangzhou Mu and
                  Jianwei Yang and
                  Jianfeng Gao and
                  Chunyuan Li and
                  Yong Jae Lee},
  title        = {{GLIGEN:} Open-Set Grounded Text-to-Image Generation},
  booktitle    = {Proceedings of {IEEE/CVF} Conference on Computer Vision and Pattern Recognition},
  pages        = {22511--22521},
  year         = {2023},
  bibsource    = {dblp computer science bibliography, https://dblp.org}
}

@INPROCEEDINGS{Avrahami_2023_CVPR,
  author={Avrahami, Omri and Hayes, Thomas and Gafni, Oran and Gupta, Sonal and Taigman, Yaniv and Parikh, Devi and Lischinski, Dani and Fried, Ohad and Yin, Xi},
  booktitle={Proceedings of IEEE/CVF Conference on Computer Vision and Pattern Recognition}, 
  title={SpaText: Spatio-Textual Representation for Controllable Image Generation}, 
  year={2023},
  volume={},
  number={},
  pages={18370-18380},}

@inproceedings{huang2023composer,
  author       = {Lianghua Huang and
                  Di Chen and
                  Yu Liu and
                  Yujun Shen and
                  Deli Zhao and
                  Jingren Zhou},
  title        = {Composer: Creative and Controllable Image Synthesis with Composable
                  Conditions},
  booktitle    = {International Conference on Machine Learning},
  series       = {Proceedings of Machine Learning Research},
  volume       = {202},
  pages        = {13753--13773},
  year         = {2023},
  bibsource    = {dblp computer science bibliography, https://dblp.org}
}

@INPROCEEDINGS{Zhang_2023_ICCV,
  author={Zhang, Lvmin and Rao, Anyi and Agrawala, Maneesh},
  booktitle={Proceedings of IEEE/CVF International Conference on Computer Vision}, 
  title={Adding Conditional Control to Text-to-Image Diffusion Models}, 
  year={2023},
  volume={},
  number={},
  pages={3813-3824},}

@inproceedings{mou2023t2i,
  title={T2i-adapter: Learning adapters to dig out more controllable ability for text-to-image diffusion models},
  author={Mou, Chong and Wang, Xintao and Xie, Liangbin and Wu, Yanze and Zhang, Jian and Qi, Zhongang and Shan, Ying},
  booktitle={Proceedings of the AAAI conference on artificial intelligence},
  volume={38},
  number={5},
  pages={4296--4304},
  year={2024}
}

@inproceedings{gal2022image,
  author       = {Rinon Gal and
                  Yuval Alaluf and
                  Yuval Atzmon and
                  Or Patashnik and
                  Amit Haim Bermano and
                  Gal Chechik and
                  Daniel Cohen{-}Or},
  title        = {An Image is Worth One Word: Personalizing Text-to-Image Generation
                  using Textual Inversion},
  booktitle    = {International Conference on Learning Representations},
  year         = {2023},
  bibsource    = {dblp computer science bibliography, https://dblp.org}
}

@inproceedings{hu2021lora,
  author       = {Edward J. Hu and
                  Yelong Shen and
                  Phillip Wallis and
                  Zeyuan Allen{-}Zhu and
                  Yuanzhi Li and
                  Shean Wang and
                  Lu Wang and
                  Weizhu Chen},
  title        = {LoRA: Low-Rank Adaptation of Large Language Models},
  booktitle    = {International Conference on Learning Representations},
  year         = {2022},
  bibsource    = {dblp computer science bibliography, https://dblp.org}
}

@INPROCEEDINGS{dai2025noisectrl,
  author={Dai, Longquan and Wang, He and Tang, Jinhui},
  booktitle={Proceedings of IEEE/CVF Conference on Computer Vision and Pattern Recognition}, 
  title={NoiseCtrl: A Sampling-Algorithm-Agnostic Conditional Generation Method for Diffusion Models}, 
  year={2025},
  volume={},
  number={},
  pages={18093-18102},
}

@article{dang2026cogblender,
  author       = {Shengqi Dang and
                  Jiaying Lei and
                  Yi He and
                  Ziqing Qian and
                  Nan Cao},
  title        = {CogBlender: Towards Continuous Cognitive Intervention in Text-to-Image
                  Generation},
  journal      = {CoRR},
  volume       = {abs/2603.09286},
  year         = {2026},
  eprinttype   = {arXiv},
  eprint       = {2603.09286},
  bibsource    = {dblp computer science bibliography, https://dblp.org}
}

@inproceedings{hertz2022prompt,
  author       = {Amir Hertz and
                  Ron Mokady and
                  Jay Tenenbaum and
                  Kfir Aberman and
                  Yael Pritch and
                  Daniel Cohen{-}Or},
  title        = {Prompt-to-Prompt Image Editing with Cross-Attention Control},
  booktitle    = {International Conference on Learning Representations},

  year         = {2023},
  bibsource    = {dblp computer science bibliography, https://dblp.org}
}

@inproceedings{hessel2021clipscore,
    title = "{CLIPS}core: A Reference-free Evaluation Metric for Image Captioning",
    author = "Hessel, Jack  and
      Holtzman, Ari  and
      Forbes, Maxwell  and
      Le Bras, Ronan  and
      Choi, Yejin",
    booktitle = "Proceedings of the Conference on Empirical Methods in Natural Language Processing",
    year = "2021",
    
pages = "7514--7528",

}

@inproceedings{wang2023exploring,
  title={Exploring clip for assessing the look and feel of images},
  author={Wang, Jianyi and Chan, Kelvin CK and Loy, Chen Change},
  booktitle={Proceedings of the AAAI conference on artificial intelligence},
  volume={37},
  number={2},
  pages={2555--2563},
  year={2023}
}
\bibliographystyle{iclr2027_conference}
\newpage

\end{document}